\documentclass[10pt,twocolumn,letterpaper]{article}

\usepackage[pagenumbers]{cvpr} 

\definecolor{cvprblue}{rgb}{0.21,0.49,0.74}
\usepackage[pagebackref,breaklinks,colorlinks,allcolors=cvprblue]{hyperref}

\usepackage{amsmath,amssymb,amsfonts}
\usepackage{algorithmic}
\usepackage{graphicx}
\usepackage{textcomp}
\usepackage{xcolor}
\usepackage{enumitem}
\usepackage{caption}
\usepackage{subcaption}
\usepackage{float}
\usepackage{longtable}
\usepackage{makecell}
\usepackage{listings}
\usepackage{adjustbox}
\usepackage{amsmath}
\usepackage{booktabs}
\usepackage{multirow}
\usepackage[hyphens]{url}
\usepackage{hyperref}
\usepackage{balance}
\usepackage[utf8]{inputenc}
\usepackage{stfloats}
\usepackage{booktabs,pifont, tabularx}
\newcommand{\cmark}{\ding{51}}  
\newcommand{\xmark}{\ding{55}}  
\newcolumntype{Y}{>{\centering\arraybackslash}X}
\usepackage{array}
\newcolumntype{L}[1]{>{\raggedright\arraybackslash}p{#1}}

\def\paperID{6124}
\def\confName{CVPR}
\def\confYear{2026}

\title{NNGPT: Rethinking AutoML with Large Language Models}

\author{
Roman Kochnev \quad Waleed Khalid \quad Tolgay Atinc Uzun \quad Xi Zhang\quad
Yashkumar S. Dhameliya \\ Furui Qin \quad Chandini Vysyaraju \quad Raghuvir Duvvuri \quad Avi Goyal \quad Dmitry Ignatov \quad Radu Timofte\\[3pt]
\small Computer Vision Lab, CAIDAS \& IFI, University of Würzburg, Germany
}
\begin{document}
\maketitle


\begin{abstract}
Building self-improving AI systems remains a fundamental challenge in the AI domain. We present NNGPT, an open-source framework that turns a large language model (LLM) into a self-improving AutoML engine for neural network development, primarily for computer vision. Unlike previous frameworks, NNGPT extends the dataset of neural networks by generating new models, enabling continuous fine-tuning of LLMs based on closed-loop system of generation, assessment, and self-improvement. It integrates within one unified workflow five synergistic LLM-based pipelines: zero-shot architecture synthesis, hyperparameter optimization (HPO), code-aware accuracy/early-stop prediction, retrieval-augmented synthesis of scope-closed PyTorch blocks (NN-RAG), and reinforcement learning. Built on the LEMUR dataset as an audited corpus with reproducible metrics, NNGPT emits from a single prompt and validates network architecture, preprocessing code, and hyperparameters, executes them end-to-end, and learns from result. The PyTorch adapter makes NNGPT framework-agnostic, enabling strong performance: NN-RAG achieves 73\% executability on 1,289 targets, 3-shot prompting boosts accuracy on common datasets, and hash-based deduplication saves hundreds of runs. One-shot prediction matches search-based AutoML, reducing the need for numerous trials. HPO on LEMUR achieves RMSE 0.60, outperforming Optuna (0.64), while the code-aware predictor reaches RMSE 0.14 with Pearson \(r=0.78\). The system has already generated over 5K validated models, proving NNGPT as an autonomous AutoML engine. Upon acceptance, the code, prompts, and checkpoints will be released for public access to enable reproducibility and facilitate community usage.

\end{abstract}

\section{Introduction}
\label{subsec:introduction}

A longstanding goal in machine learning is to build systems that design and improve themselves. In practice, assembling a competitive neural pipeline still requires substantial expert effort: choosing architectures, writing transforms, specifying losses and metrics, tuning hyperparameters, and running many training jobs. Classical AutoML, for example Optuna~\cite{10.1145/3292500.3330701} with Tree-structured Parzen Estimator (TPE)~\cite{10.5555/2986459.2986743}, automates parts of this process but relies on black-box search that requires hundreds of trials and ignores the semantics of the underlying code.

Large language models (LLMs) such as GPT-4~\cite{openai2024gpt4technicalreport}, Code Llama~\cite{rozière2024codellamaopenfoundation}, and DeepSeek~\cite{deepseekai2025deepseekv3technicalreport} excel at code generation, which has motivated attempts to let them propose architectures, hyperparameters, or training scripts. Most prior work, however, remains at design time: the LLM emits fragments or templates that are only loosely tied to execution, logging, or learning from outcomes, so the feedback loop from generation to training to improved generation is rarely closed.

We ask whether a single LLM, embedded in an online loop with executable code and training feedback, can serve as the core of a self-improving AutoML engine for neural networks. Answering this requires representing full training pipelines as executable specifications, validating and running them at scale, predicting performance early, and using logs to continuously improve the LLM.

We introduce \textbf{NNGPT}, an open-source framework that, to our knowledge, is the first to unify zero-shot model generation and editing, hyperparameter recommendation, accuracy and early-stop prediction, retrieval-augmented code synthesis, and an reinforcement learning (RL)-based improvement loop into a single closed AutoML cycle driven by prompt-based LLM inference. From one high-level prompt, NNGPT produces an executable training specification (model, transforms, metrics, optimizer, schedule), executes it, logs code and metrics, and uses these logs to fine-tune and reinforce the underlying LLMs.

\begin{table*}[t]
\centering
\scriptsize               
\setlength{\tabcolsep}{4pt}  
\renewcommand{\arraystretch}{0.92} 
\begin{tabular}{lcccccccc}
\toprule
\textbf{Method} & \textbf{LLM} & \textbf{Arch.} & \textbf{HPO} & \textbf{Full train} & \textbf{One-shot} & \textbf{Closed loop} & \textbf{Acc. pred.} & \textbf{Open source} \\
\midrule
Optuna~\cite{10.1145/3292500.3330701}       & \xmark & \xmark & \cmark & \xmark & \xmark & \xmark & \xmark & \cmark \\
Hyperopt~\cite{bergstra2013hyperopt}        & \xmark & \xmark & \cmark & \xmark & \xmark & \xmark & \xmark & \cmark \\
Google Vizier~\cite{golovin2017googlevizier}& \xmark & \xmark & \cmark & \xmark & \xmark & \xmark & \xmark & \cmark \\
\midrule
ENAS~\cite{pham2018efficient}               & \xmark & \cmark & \cmark & \xmark & \xmark & \xmark & \xmark & \cmark \\
DARTS~\cite{liu2018darts}                   & \xmark & \cmark & \cmark & \xmark & \xmark & \xmark & \xmark & \cmark \\
AutoKeras~\cite{jin2019auto}                & \xmark & \cmark & \cmark & \xmark & \xmark & \xmark & \xmark & \cmark \\
\midrule
EvoPrompting~\cite{chen2023evoprompting}    & \cmark & \cmark & \cmark & \xmark & \xmark & \cmark & \xmark & \xmark \\
LLAMBO~\cite{liu2024largelanguagemodelsenhance} & \cmark & \xmark & \cmark & \xmark & \xmark & \cmark & \xmark & \xmark \\
LLMatic~\cite{nasir2024llmatic}             & \cmark & \cmark & \cmark & \xmark & \xmark & \cmark & \xmark & \xmark \\
GPT-NAS~\cite{yu2023gptnas}                 & \cmark & \cmark & \cmark & \xmark & \xmark & \cmark & \xmark & \xmark \\
Self-Programming AI~\cite{sheng2023selfprogramming} & \cmark & \cmark & \cmark & \xmark & \xmark & \cmark & \xmark & \xmark \\
Grammar / schema prompting~\cite{wang2023grammar} & \cmark & \xmark & \xmark & \xmark & \cmark & \xmark & \xmark & \xmark \\
\midrule
\textbf{NNGPT (ours)}                       & \cmark & \cmark & \cmark & \cmark & \cmark & \cmark & \cmark & \cmark \\
\bottomrule
\end{tabular}
\caption{Feature comparison between NNGPT and representative AutoML and LLM-based methods. Columns indicate whether a method uses an LLM, generates architectures (Arch.), performs hyperparameter optimization (HPO), synthesizes full training specifications, operates in a one-shot fashion, supports a closed feedback loop, includes accuracy prediction, and is open source.}
\label{tab:method_comparison}
\end{table*}

NNGPT is built on LEMUR~\cite{goodarzi2025lemurneuralnetworkdataset}, which we treat as a corpus of executable neural programs with audited implementations, unified preprocessing, and reproducible metrics. This grounds one-shot generation in runnable PyTorch~\cite{paszke2019pytorchimperativestylehighperformance} code rather than abstract templates. The framework remains PyTorch-agnostic via a thin adapter that exposes model factories, data transforms, and metric bindings. In practice, NNGPT has generated and trained over 10,000 distinct models (over 5,000 through its self-improving loop with LLMs), with training statistics, all incorporated into LEMUR as verified outcomes.

The closed-loop view is as follows. A fine-tuned LLM emits network architectures, preprocessing code, and hyperparameters; an execution engine compiles, trains, and logs; a code-aware predictor forecasts final accuracy and a stopping epoch from model code plus early metrics; a retrieval module patches or synthesizes PyTorch blocks; and an RL layer updates LoRA adapters~\cite{lora} using rewards derived from executability and downstream performance. 

Our experiments demonstrate that LLMs can function as autonomous AutoML systems: they generate complete pipelines, predict performance, and refine themselves based on real-time feedback within a continuous self-improvement loop.

\subsection{Related Work}
\label{subsec:related_work}

Prior work falls into two strands: (i) search-based AutoML/NAS and (ii) LLM-driven generation.

\textbf{Search-based AutoML and NAS.}
Classical HPO frameworks such as Optuna~\cite{10.1145/3292500.3330701}, Hyperopt~\cite{bergstra2013hyperopt}, and Google Vizier~\cite{golovin2017googlevizier} rely on Bayesian optimization or TPE to iteratively probe the space of hyperparameters. These tools are strong baselines but typically require hundreds of trials, treat training code as an opaque black box, and do not synthesize \emph{complete} train specs (data, metrics, optimizer). NAS systems, including ENAS~\cite{pham2018efficient}, DARTS~\cite{liu2018darts}, and AutoKeras~\cite{jin2019auto}, extend search to architectures (with many stabilized variants), yet remain compute-intensive and slow to adapt to new tasks and codebases.

\textbf{LLM-driven generation and closed loops.}
Recent methods integrate LLMs into design loops: EvoPrompting~\cite{chen2023evoprompting} uses an LLM as a mutation operator in evolutionary search; LLAMBO~\cite{liu2024large} couples LLM suggestions with Bayesian optimization; LLMatic~\cite{nasir2024llmatic}, GPT-NAS~\cite{yu2023gptnas}, and Self-Programming AI~\cite{sheng2023selfprogramming} generate code or fill templates. Grammar/schema prompting~\cite{wang2023grammar} and retrieval help with validity. However, most systems either require repeated LLM queries, cover only fragments of the pipeline (e.g., mutations without training integration), or lack an online mechanism to learn from executed runs.

\textbf{Performance prediction.}
Predicting final accuracy and early stopping has been studied via curve fitting from early learning curves~\cite{domhan2015, adriaensen2023}, architecture- or text-based surrogates~\cite{long2020, white2021}, and graph-level representations~\cite{ding2024}. These approaches often abstract away implementation details (losses, custom layers, metric code) that strongly affect convergence, and they rarely integrate with executable pipelines.

\textbf{Position of NNGPT.}
Table~\ref{tab:method_comparison} situates NNGPT among these lines. Unlike prior art, NNGPT (1) \emph{one-shot} generates schema-validated \emph{full} training specifications (architecture \& HPO) that run immediately; (2) replaces costly search for many tasks with LLM-based hyperparameter recommendation; (3) includes a \emph{code-aware} predictor that consumes executable code and early metrics to forecast final accuracy and a stopping epoch; and (4) closes the loop via LoRA/RL updates using execution logs - within a single open-source framework. A comprehensive literature review with extended comparisons is provided in the Supplementary Material, Sec.~\ref{section:extended_related_work}.

\begin{figure*}
    \centering
    \includegraphics[width=1\linewidth]{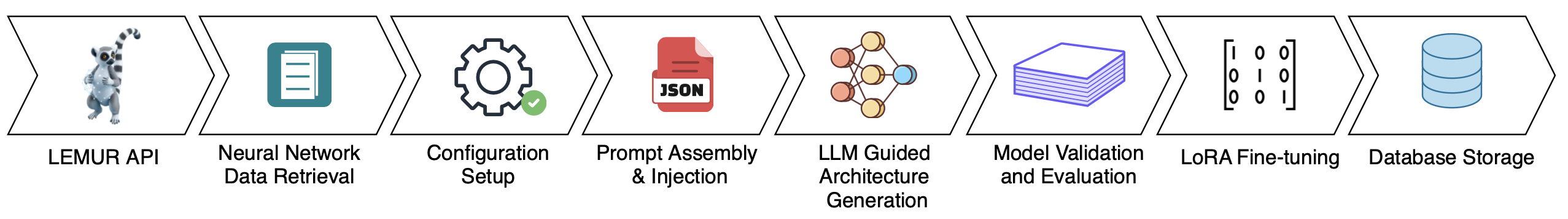}
    \caption{Overview of the NNGPT Pipeline: starting with a query to the LEMUR API to retrieve a neural network entry and its metadata, the system constructs a prompt, using an LLM generates the code of neural network model, and trains it. Artifacts are logged in a structured database, while LoRA fine-tuning continuously updates the LLM based on training results, creating a self-improving AutoML loop.}  
    \label{figure:pipeline_main_all_parts}
\end{figure*}

\section{NNGPT Framework}
\label{sec:nngpt-framework}

\subsection{System Overview}
\label{subsec:system-overview}
NNGPT exposes LLMs as end to end configuration agents for neural networks. Given a high level task description, the system constructs a structured prompt, calls an LLM once, validates the structured output, executes the resulting training pipeline, logs all artefacts, and feeds the trace back into its training data.

At the core of NNGPT is a shared execution substrate built on the LEMUR dataset framework, which provides executable PyTorch models, standardized training loops, and reproducible evaluation metrics. To avoid hard coupling to LEMUR, the framework defines a thin adapter interface for arbitrary PyTorch codebases that maps a configuration object to a model constructor, data transforms, and metric definitions. Any model that implements this interface can be driven by the same LLM generated specification.

A typical pass through the system proceeds as follows. First, the configuration module aggregates task level information such as dataset, objective, resource limits, and optional seed architectures. Next, a prompt builder converts this state into a JSON like instruction block that encodes the desired output schema. The LLM backend then emits a structured, machine readable specification with explicit fields for architecture, hyperparameters, and training procedure. A schema layer based on Pydantic validates the output and either autocorrects minor issues or requests a single regeneration in case of structural violations. Valid configurations are handed off to a distributed training engine, which runs the experiment, logs metrics and metadata in an SQLite store, and optionally exports models back into the LEMUR corpus. Finally, dedicated modules use these traces to fine tune the LLM with LoRA, to train an accuracy and early stopping predictor, or to update a RL signal. This shared infrastructure underpins all pipelines described below.


\begin{figure*}[t]
  \centering
  \includegraphics[width=\textwidth]{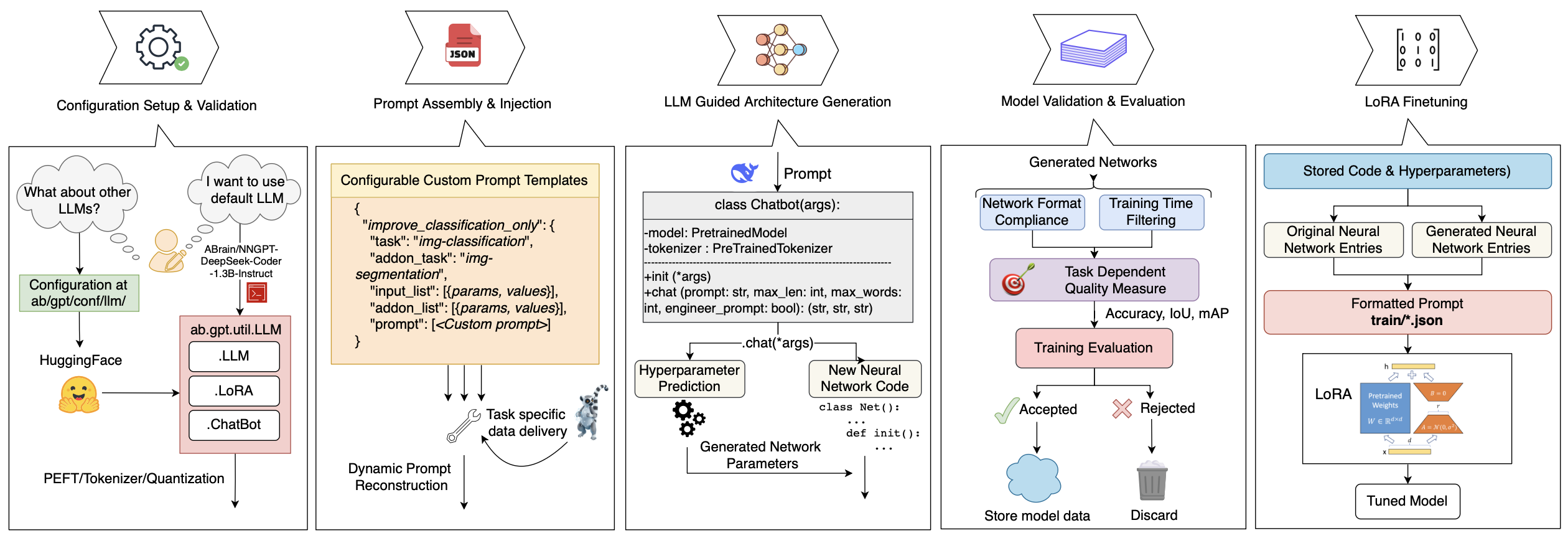}
  \caption{Detailed view of pipeline stages 3–7 in NNGPT. Left to right: Configuration Setup \& Validation (default or HF-loaded LLM, PEFT/quantization); Prompt Assembly \& Injection (templated prompts with task data); LLM-Guided Architecture Generation (one-shot code and hyperparameters); Model Validation \& Evaluation (format checks, training, logging); LoRA Fine-tuning (update adapters from \texttt{train/*.json}). All pipelines reuse this stack; variation mainly occurs in the Prompt stage.}
  \label{figure:pipeline_parts_3-7}
\end{figure*}


\subsection{Problem Formulation}
\label{subsec:problem-formulation}

We cast NNGPT as closed-loop neural pipeline synthesis problem. Given a task prompt \(P\) (task, dataset, metric, optional budget) and an optional retrieval set \(R\) (code and metadata from LEMUR or external projects), the goal is to produce an executable training specification \(C\) and a horizon \(T\) that maximize downstream performance under a compute budget.

An LLM generator \(G_{\theta}\) maps
\begingroup
\setlength{\abovedisplayskip}{6pt}%
\setlength{\belowdisplayskip}{6pt}%
\setlength{\abovedisplayshortskip}{6pt}%
\setlength{\belowdisplayshortskip}{6pt}%
\[
(P,R)\;\mapsto\;\tilde{C}\in\mathcal{C},
\]
\endgroup
where \(\tilde{C}\) is a YAML/JSON spec that defines the model, data transforms, optimization schedule, loss, and nominal epochs. A validator \(V\) (schema + type/shape checks) enforces structural constraints, auto-fixes minor issues or triggers a single re-prompt; only \(V(\tilde{C})=\mathrm{pass}\) is promoted to \(C\).

Given \(C\) and \(T\), the trainer \(E\) executes the pipeline,
\begingroup
\setlength{\abovedisplayskip}{6pt}%
\setlength{\belowdisplayskip}{6pt}%
\setlength{\abovedisplayshortskip}{6pt}%
\setlength{\belowdisplayshortskip}{6pt}%
\[
(\mathbf{m},\mathbf{u}) = E(C,T),
\]
\endgroup
producing validation metrics over epochs \(\mathbf{m}\) and auxiliary logs \(\mathbf{u}\) (loss curves, learning rate, runtime, hardware). Code and traces are persisted in a structured log.

To cut wasted compute, a code-aware predictor \(H_{\phi}\) estimates
\begingroup
\setlength{\abovedisplayskip}{6pt}%
\setlength{\belowdisplayskip}{6pt}%
\setlength{\abovedisplayshortskip}{6pt}%
\setlength{\belowdisplayshortskip}{6pt}%
\[
(\widehat{\mathrm{acc}}_{\!*},\widehat{t}_{\!*}) = H_{\phi}(C,R,\mathbf{m}_{1:t_0}),
\]
\endgroup
where \(\widehat{\mathrm{acc}}_{\!*}\) approximates the final accuracy and \(\widehat{t}_{\!*}\) proposes an early-stopping epoch from the first \(t_0\) epochs. This induces an adaptive horizon \(T'=\widehat{t}_{\!*}\) and enables early termination and priority scheduling.

Over time the log
\begingroup
\setlength{\abovedisplayskip}{6pt}%
\setlength{\belowdisplayskip}{6pt}%
\setlength{\abovedisplayshortskip}{6pt}%
\setlength{\belowdisplayshortskip}{6pt}%
\[
\mathcal{D}_{\text{log}}=\{(P^{(k)},R^{(k)},C^{(k)},\mathbf{m}^{(k)},\mathbf{u}^{(k)})\}_{k=1}^{K}
\]
\endgroup
updates \(\theta\) and \(\phi\): \(G_{\theta}\) is refined with LoRA fine-tuning and lightweight policy-gradient updates using rewards from executability and downstream performance, while \(H_{\phi}\) is trained to minimize prediction error on final accuracy and stopping epoch. This closes the loop: generate, validate, execute, log, update, under an explicit compute budget.

\subsection{Pipelines in NNGPT}
\label{subsec:pipelines}

On top of the common backbone, NNGPT implements five concrete pipelines that target different stages of the model development process while reusing the same prompting, validation, and execution stack. As summarized in Fig.~\ref{figure:pipeline_main_all_parts}, all pipelines follow one seven-step template: LEMUR retrieval $\rightarrow$ configuration setup/validation $\rightarrow$ prompt assembly \& injection $\rightarrow$ LLM-guided generation $\rightarrow$ model validation/training $\rightarrow$ LoRA fine-tuning $\rightarrow$ database logging. The pipeline-specific behavior is concentrated in the \emph{Prompt} stage (steps 3–4): the prompt schema, retrieved context, and expected outputs differ across zero-shot generation/editing, hyperparameter recommendation, accuracy/early-stop prediction, NN-RAG patching, and the RL loop. Implementation details of the shared stages are expanded in Fig.~\ref{figure:pipeline_parts_3-7}.


\subsubsection{Zero-shot model generation and editing}
\label{subsubsec:zero-shot-generation}

The first pipeline turns LLMs into generators of complete architectures and training specifications. Given a task prompt and a target dataset, NNGPT queries LEMUR for high-performing models, selects one \emph{reference} implementation and $n$ additional \emph{supporting} examples, and assembles a Few-Shot Architecture Prompting (FSAP) prompt that concatenates the task description, dataset specs, and full PyTorch classes for these models. The LLM (typically a LoRA-fine-tuned DeepSeek-Coder-7B) is instructed to emit a new scope-closed architecture class plus a YAML configuration rather than a local patch. A whitespace-normalized MD5 hash validator removes duplicate programs in $<1$\,ms, preventing redundant training of architectures that differ only in formatting. Generated specifications are checked by the schema layer and executed by the training engine; successful runs are logged and added back to LEMUR as new entries. See Sec.~\ref{subsec:zero-shot_generation} for the experimental setup and evaluation.

\subsubsection{Hyperparameter recommendation}
\label{subsubsec:hyperparameter_recommendation}

The second pipeline treats hyperparameter tuning as conditional text generation over a fixed schema. Training data are drawn from LEMUR runs that link each model to its dataset, transforms, target accuracy, epoch budget, and the hyperparameters that achieved this performance. Prompts expose the model name, \texttt{nn\_code}, \texttt{transform\_code}, task, dataset, and either a target accuracy or a “maximize accuracy’’ instruction; the LLM is asked to output only the values for a predefined set of hyperparameters (learning rate, momentum, weight decay, batch size, etc.) in a strict order.\ 

We fine-tune several DeepSeek and CodeLlama variants with LoRA on this corpus and integrate the best checkpoints (DeepSeek-Coder-1.3B-Base and DeepSeek-R1-Distill-Qwen-7B) as plug-and-play recommenders inside NNGPT. At inference time, the hyperparameter pipeline runs as a one-shot replacement for search-based HPO: a single LLM call produces a configuration that is executed by the same backend as Optuna, and outcomes are logged back to LEMUR. Section~\ref{subsec:hyperparameter_generation_quality} shows that these one-shot recommendations reach accuracy comparable to TPE-based Optuna on LEMUR, while eliminating thousands of optimization trials.


\subsubsection{Accuracy and early-stop prediction}
\label{subsubsec:accuracy_early_stop_prediction}

The third pipeline uses an LLM as a multi-task predictor of training dynamics. Each training run in LEMUR is converted into a structured prompt that fuses three modalities: (i) structured metadata (task, dataset, maximum epochs), (ii) executable code snippets for the model, metric, and data transforms, and (iii) early-epoch validation accuracies (typically at epochs 1 and 2).\ A DeepSeek-Coder-1.3B backbone is adapted with 4-bit QLoRA, reducing memory by $\sim$4$\times$ while keeping full context, and fine-tuned to regress both the final best validation accuracy and the epoch at which it occurs under a constrained JSON/YAML output schema.\ Stratified group splitting by (task, dataset, architecture) prevents leakage between similar models and enforces out-of-architecture generalization.

Within NNGPT, the predictor runs alongside training: after a few epochs, it estimates $(\widehat{\mathrm{acc}}_{\!*},\widehat{t}_{\!*})$ and can trigger early stopping when predicted accuracy is low, or prioritize promising runs in the job scheduler. As shown in Sec.~\ref{subsec:accuracy-prediction}, the baseline model achieves RMSE $\approx 0.145$ and Pearson $r\!\approx\!0.78$ over 1{,}200+ runs, making it practical for compute-aware filtering and scheduling.


\subsubsection{NN-RAG: retrieval augmented patching}
\label{subsubsec:nn-rag}
The fourth pipeline augments generation with a retrieval system for reusable PyTorch blocks. NN-RAG scans curated repositories for non-abstract \texttt{nn.Module} subclasses that implement \texttt{forward()}, round-trips their source through LibCST~\cite{libcst} to preserve concrete syntax, and indexes parse trees and import relations in an SQLite-backed store with SHA-1 content addressing.\ Given a target architecture and a textual or code query (e.g., “multi-head attention block’’), NNGPT retrieves candidate blocks and uses a scope-sensitive resolver to compute the minimal transitive closure of dependencies that respects Python’s LEGB and import semantics. Definitions are reordered via topological sort to satisfy definition-before-use constraints, and the resulting scope-closed module is passed through AST parsing, bytecode compilation, and sandboxed import. On a corpus of 1{,}289 target blocks, this validator admits 941 modules (73\% executability), covering attention, convolutions, transformer blocks~\cite{vaswani2023attentionneed}, losses, and higher-level architectures. These validated blocks form a library that NNGPT can splice into generated models, enabling retrieval-augmented patching while preserving executability guarantees.


\subsubsection{Reinforcement learning based closed loop}
\label{subsubsec:rl_closed_loop}
We add an RL layer on top of supervised fine-tuning, treating the LLM as a policy over architecture code and configurations. From LEMUR we build a masked-architecture corpus where layer definitions (e.g., \texttt{self.conv1 = nn.Conv2d(...)}) are replaced by placeholders; the model learns to reconstruct these blocks, acquiring realistic priors. At RL time it proposes full architectures that are executed via \texttt{check\_nn}, which tests compilation, forward correctness, and a short CIFAR-10~\cite{cifar-10} run. A scalar reward mixes syntactic validity, runtime feasibility, and validation accuracy, and a GRPO-style policy-gradient update (inspired by DeepSeek-R1~\cite{deepseekai2025deepseekr1incentivizingreasoningcapability}) is applied to LoRA adapters.


In parallel, a channel-mutation engine performs deterministic width edits through five steps: \texttt{torch.fx} tracing, mapping graph nodes to source, planning shape-preserving changes, AST rewriting, and re-validation. Both RL-generated and mutated models pass through the same validation and training stack, and successful configurations are added to LEMUR, closing a performance-aware loop that explores and reinforces architectures that train well in practice.



Across all five pipelines, accepted configurations, logs, and artifacts are exported back into LEMUR dataset or external repositories, turning raw LLM generations into a continually expanding corpus of executable neural programs that subsequent NNGPT runs can query, modify, and build upon.

\section{Experiments}
\label{sec:experiments}

Experiments are conducted on a single 24GB GPU (Geforce RTX 3090/4090) at a time, using PyTorch and 8-bit or 4-bit LoRA adapters on top of code-capable LLMs within the AI Linux\footnote{AI Linux: \scriptsize \url{https://hub.docker.com/r/abrainone/ai-linux}} environment on the Kubernetes cluster. NNGPT interacts with the LEMUR corpus through a fixed API that exposes model factories, transformation pipelines, and metric definitions, but any PyTorch codebase can be integrated via the same adapter interface. For each pipeline we report results on mid-scale vision benchmarks (e.g.\ CIFAR-like classification and COCO-style detection/segmentation tasks), using standardized training and evaluation protocols from LEMUR. Further engineering details, CLI tools, and prompts are provided in the Supplementary Material, Sec.~\ref{section:implementation}.

\subsection{Zero-Shot Generation and LEMUR Growth}
\label{subsec:zero-shot_generation}

To quantify NNGPT's ability to expand LEMUR via zero-shot generation, we instantiate a Few-Shot Architecture Prompting (FSAP) pipeline. For each base model and dataset, the LLM receives a natural-language task description, one reference implementation, and $n \in \{1,\dots,6\}$ supporting architectures sampled from strong LEMUR entries, and is asked to synthesize a new PyTorch model that follows the required API. We denote the resulting variants \texttt{alt-nn1}–\texttt{alt-nn6}. Generated code passes through schema validation and a whitespace-normalized MD5 hash check to remove duplicates before training.

Table~\ref{table:nn_gen_counts} summarizes generation statistics. With $n=1$ (\texttt{alt-nn1}), NNGPT produces 3{,}394 valid architectures; increasing the number of supporting examples sharply reduces success, down to 306 models for $n=2$ and $\approx 100$ models for $n=3$–$5$. At $n=6$ (\texttt{alt-nn6}) the system generates only 7 models (99.8\% failure), indicating severe context overflow. Across all settings, we obtain 4{,}033 candidates, of which about 1{,}900 are unique after hash-based deduplication; the hash validator rejects $\sim$100 near-duplicate programs and saves an estimated 200–300 GPU hours of redundant training.

\begin{table}[t]
    \centering
    \small
    \setlength{\tabcolsep}{5pt}
    \begin{tabular}{lccc}
        \toprule
        \textbf{Variant} & \textbf{$n$ (examples)} & \textbf{Models} & \textbf{Success rate} \\
        \midrule
        \texttt{alt-nn1} & 1 & 3{,}394 & 100\% \\
        \texttt{alt-nn2} & 2 & 306     & 9.0\% \\
        \texttt{alt-nn3} & 3 & 103     & 3.0\% \\
        \texttt{alt-nn4} & 4 & 102     & 3.0\% \\
        \texttt{alt-nn5} & 5 & 121     & 3.6\% \\
        \texttt{alt-nn6} & 6 & 7       & 0.2\%\textsuperscript{\dag} \\
        \midrule
        \textbf{Total}   & -- & \textbf{4{,}033} & -- \\
        \bottomrule
    \end{tabular}
    \caption{\textbf{FSAP yield vs.\ context size.} The LLM gets one reference model and $n$ supporting examples from LEMUR and must emit a complete, executable PyTorch architecture. Variants \texttt{alt-nn$k$} use $k$ supports. \textit{Models} count schema-validated, deduplicated architectures; \textit{Success rate} is yield relative to $n{=}1$ (100\%). The drop at $n{=}6$ indicates context overflow. In total, 4{,}033 candidates were produced ($\sim$1.9k unique; $\sim$100 near-duplicates filtered).}
    \label{table:nn_gen_counts}
    \vspace{0.25em}
    {\footnotesize\textsuperscript{\dag}\,99.8\% relative drop vs.\ $n{=}1$ indicates severe context overflow.}
\end{table}

\begin{table}[t]
    \centering
    \small
    \setlength{\tabcolsep}{6pt}
    \begin{tabular}{lccc}
        \toprule
        \textbf{Variant} & \textbf{$n$} & \textbf{Models} & \textbf{Balanced Mean (\%)} \\
        \midrule
        \texttt{alt-nn1} & 1 & 3{,}394 & $51.5 \pm 29.5$ \\
        \texttt{alt-nn2} & 2 & 306     & $49.8 \pm 32.3$ \\
        \textbf{\texttt{alt-nn3}} & \textbf{3} & \textbf{103} & \textbf{$53.1 \pm 26.9$} \\
        \texttt{alt-nn4} & 4 & 102     & $47.3 \pm 32.5$ \\
        \texttt{alt-nn5} & 5 & 121     & $43.0 \pm 33.1$ \\
        \bottomrule
    \end{tabular}
        \caption{\textbf{Accuracy of generated architectures (dataset-balanced mean).} For each variant \texttt{alt-nn$k$} we train every accepted model for one epoch with SGD+momentum and report top-1 accuracy. To avoid bias from uneven per-dataset sample counts, we first average within each (variant, dataset) pair and then macro-average across datasets (equal weight per dataset). \texttt{alt-nn6} is omitted due to insufficient sample size ($n{=}7$). \textbf{Bold} marks the best overall balanced mean.}
    \label{table:nn_gen_performance}
\end{table}

To compare quality across datasets with different sample sizes, we report dataset-balanced means (Table~\ref{table:nn_gen_performance}). The balanced mean accuracy peaks at $n=3$ (\texttt{alt-nn3}, 53.1\%) compared to 51.5\% for the $n=1$ baseline, while larger contexts ($n=4,5$) degrade performance to 47.3\% and 43.0\%. Gains are most pronounced on harder tasks such as CIFAR-100, where \texttt{alt-nn3} improves over \texttt{alt-nn1} by +11.6 percentage points after a single epoch. In aggregate, this pipeline contributes roughly 1.9k new architectures (and over 3k trained runs) to LEMUR, demonstrating that NNGPT can continuously grow its own training corpus through LLM-driven zero-shot generation.

\begin{table}[t]
\scriptsize
\centering
\setlength{\tabcolsep}{2.5pt}
\begin{tabular}{r L{0.46\linewidth} c c c c}
\toprule
\textbf{Exp.} & \textbf{Model} & \textbf{Ep.} & \makecell{\textbf{Correct}\\\textbf{/500}} & \textbf{Failed} & \makecell{\textbf{Pct.}\\(\%)}\\
\midrule
1  & DeepSeek-Coder-V2-Lite-Base              & 15 &  8  & 492 &  1.60 \\
\hline
2  & DeepSeek-Coder-1.3b-base                 & 10 & 435 &  65 & 87.00 \\
\textbf{3}  & \textbf{DeepSeek-Coder-1.3b-base}        & \textbf{15} & \textbf{442} & \textbf{58}  & \textbf{88.40} \\
4  & DeepSeek-Coder-1.3b-base                 & 25 &  0  & 500 &  0.00 \\
5  & DeepSeek-Coder-1.3b-base                 & 35 &  4  & 496 &  0.80 \\
\hline
6  & DeepSeek-R1-Distill-Qwen-7B              & 15 & 457 &  43 & 91.40 \\
\textbf{7}  & \textbf{DeepSeek-R1-Distill-Qwen-7B}     & \textbf{20} & \textbf{465} & \textbf{35}  & \textbf{93.00} \\
8  & DeepSeek-R1-Distill-Qwen-7B              & 35 & 176 & 324 & 35.20 \\
9  & DeepSeek-R1-Distill-Qwen-7B              & 35 & 268 & 232 & 53.60 \\
\hline
10 & DeepSeek-Coder-7b-base-v1.5              & 15 & 127 & 373 & 25.40 \\
11 & DeepSeek-Coder-7b-base-v1.5              & 20 & 114 & 386 & 22.80 \\
12 & DeepSeek-Coder-7b-base-v1.5              & 35 &  0  & 500 &  0.00 \\
\hline
13 & DeepSeek-Math-7b-base                    & 15 & 214 & 286 & 42.80 \\
14 & DeepSeek-Math-7b-base                    & 35 &  1  & 499 &  0.20 \\
\hline
\textbf{15} & \textbf{DeepSeek-Coder-7b-Instruct-v1.5} & 15 & 368 & 132 & 73.60 \\
16 & DeepSeek-Coder-7b-Instruct-v1.5          & 20 & 368 & 132 & 73.60 \\
17 & DeepSeek-Coder-7b-Instruct-v1.5          & 35 &  23 & 477 &  4.60 \\
\hline
\textbf{18} & \textbf{CodeLlama-7b-Python-hf}          & \textbf{21} & \textbf{460} &  40 & \textbf{92.00} \\
\bottomrule
\end{tabular}
\caption{\textbf{Structured hyperparameter generation on 500 held-out LEMUR configs.} Each row is a fine-tuned checkpoint (model, epochs). For every config, the LLM must output a complete hyperparameter set that matches a strict schema; a prediction is counted as \emph{Correct} if it passes all schema/value checks (else \emph{Failed}). Percentage is Correct/500. Bold = best per family and overall.}
\label{table:hyperparams_response_quality}
\end{table}

\subsection{Hyperparameter Generation Quality}
\label{subsec:hyperparameter_generation_quality}

We evaluate how well fine-tuned DeepSeek and CodeLlama models generate \emph{valid} hyperparameter sets from structured prompts. Given a specification describing model architecture, dataset, task, transform pipeline, and desired accuracy, the LLM must output a complete hyperparameter configuration (names and values) that conforms to the expected schema (Listing~\ref{listing:generation_prompt_format}). Evaluation is carried out on 500 held-out neural configurations sampled from LEMUR; a generation is counted as correct if it passes schema and value-type checks.

\begin{table}[t]
\scriptsize
\centering
\setlength{\tabcolsep}{3pt}
\begin{tabular}{l c c c}
\toprule
\multicolumn{4}{c}{\textbf{17 CV Models (as in \cite{kochnev2025optunavscodellama})}}\\
\midrule
\textbf{Method} & \textbf{RMSE} & \textbf{SE} & \textbf{95\% CI} \\
\midrule
Optuna$^{*}$                 & 0.589 & 0.004 & [0.581, 0.597] \\
CodeLlama-7b-Python-hf$^{*}$ & \textbf{0.563} & 0.004 & [0.556, 0.570] \\
DeepSeek-Coder-1.3b-base     & 0.651 & 0.032 & [0.588, 0.714] \\
DeepSeek-R1-Distill-Qwen-7B  & 0.659 & 0.037 & [0.586, 0.732] \\
CodeLlama-7b-Python-hf       & 0.642 & 0.031 & [0.554, 0.729] \\
\midrule
\multicolumn{4}{c}{\textbf{LEMUR Dataset}}\\
\midrule
\textbf{Method} & \textbf{RMSE} & \textbf{SE} & \textbf{95\% CI} \\
\midrule
Optuna                        & 0.636 & 0.002 & [0.633, 0.640] \\
DeepSeek-Coder-1.3b-base      & 0.649 & 0.062 & [0.526, 0.771] \\
DeepSeek-R1-Distill-Qwen-7B   & 0.652 & 0.029 & [0.595, 0.709] \\
CodeLlama-7b-Python-hf        & \textbf{0.603} & 0.031 & [0.543, 0.663] \\
\bottomrule
\end{tabular}
\caption{\textbf{Downstream training with LLM-generated hyperparameters vs.\ Optuna.} We compare one-shot LLM recommendations (no iterative search; 35–50 predictions per model) against Optuna (${>}20$k trials) on two evaluation sets. Metric: RMSE↓ of $\epsilon_i{=}1{-}a_i$, where $a_i$ is the observed validation accuracy when trained with predicted HPs. We report SE and normal-approx.\ 95\% CI. Rows $^{*}$ reproduced from~\cite{kochnev2025optunavscodellama}.}
\label{table:hyperparams_RMSE_std_conf_intervals}
\end{table}

Table~\ref{table:hyperparams_response_quality} reports the number of correct generations. The best overall result is obtained by DeepSeek-R1-Distill-Qwen-7B~\cite{deepseekai2025deepseekr1incentivizingreasoningcapability} (20 epochs), with 465 valid outputs (93.00\%), slightly outperforming our fine-tuned CodeLlama-7b-Python-hf~\cite{rozière2024codellamaopenfoundation} (460/500, 92.00\%), which has previously been used as a strong baseline for this task~\cite{kochnev2025optunavscodellama}. Smaller and domain-specialized DeepSeek variants also perform well: DeepSeek-Coder-1.3b-base~\cite{guo2024deepseekcoderlargelanguagemodel} reaches 88.40\% after 15 epochs, and DeepSeek-Coder-7b-Instruct-v1.5~\cite{guo2024deepseekcoderlargelanguagemodel} achieves a stable 73.60\% at 15–20 epochs. At the same time, several runs with prolonged fine-tuning collapse to near-zero valid generations (e.g., 25–35 epochs for DeepSeek-Coder-1.3b-base and DeepSeek-Math-7b-base~\cite{shao2024deepseekmathpushinglimitsmathematical}), highlighting the sensitivity of structured generation tasks to overfitting and over-training.

To assess downstream effectiveness, we use the generated hyperparameters to train models on two evaluation sets: (i) the 17-model CV benchmark from~\cite{kochnev2025optunavscodellama} and (ii) the LEMUR dataset. For each configuration, we compute an error term \(\epsilon_i = 1 - a_i\), where \(a_i\) is the observed accuracy under the predicted hyperparameters, and report RMSE over \(\epsilon_i\) along with standard errors and 95\% confidence intervals (Table~\ref{table:hyperparams_RMSE_std_conf_intervals}).

On the 17-model benchmark, prior results for CodeLlama-7b-Python-hf~\cite{kochnev2025optunavscodellama} achieve an RMSE of 0.563, statistically outperforming Optuna~\cite{akiba2019optuna} (0.589). Our DeepSeek-based models yield higher RMSEs and wider intervals on this heterogeneous testbed, consistent with their sensitivity to overfitting. On the LEMUR distribution, where schemas and transforms align more closely with training, DeepSeek-Coder-1.3b-base and DeepSeek-R1-Distill-Qwen-7B obtain RMSEs of 0.649 and 0.652, close to Optuna’s 0.636, while our fine-tuned CodeLlama-7b-Python-hf achieves the best RMSE of 0.603. Although LLM-based methods show wider confidence intervals (partly because each model is evaluated on only 35–50 one-shot predictions compared with over 20\,000 trials for Optuna), they approach or exceed Optuna’s accuracy while avoiding iterative search, demonstrating that one-shot LLM hyperparameter generation can serve as a competitive, low-cost alternative in the NNGPT loop.

\begin{table}[t]
\scriptsize
\centering
\setlength{\tabcolsep}{1pt}
\begin{tabular}{l c c L{0.15\linewidth}}
\toprule
\textbf{Dataset} & \textbf{RMSE [95\% CI]} & \textbf{Pearson $r$ (p)} & \textbf{Notes} \\
\midrule
CelebA-Gender & 0.0497 [0.044, 0.056] & 0.2125 ($p=2.38\times10^{-7}$)   & Low correlation despite low RMSE \\
CIFAR-10      & 0.1589 [0.150, 0.168] & 0.3472 ($p=3.77\times10^{-30}$) & Moderate correlation \\
CIFAR-100     & 0.1668 [0.155, 0.178] & $-0.0276$ (\textit{ns})         & Very poor ranking ability \\
ImageNette    & 0.2439 [0.221, 0.266] & 0.4973 ($p=1.25\times10^{-36}$) & High error, moderate $r$ \\
MNIST         & 0.0331 [0.025, 0.040] & 0.4224 ($p=1.04\times10^{-22}$) & Excellent precision \\
Places365     & 0.1672 [0.135, 0.193] & $-0.0297$ (\textit{ns})         & Weak generalization \\
SVHN          & 0.0605 [0.058, 0.063] & 0.3587 ($p=4.08\times10^{-20}$) & Low $r$, tolerable RMSE \\
COCO          & 0.1494 [0.138, 0.160] & 0.8173 ($p<10^{-4}$)            & Strongest dataset correlation \\
\midrule
\textbf{Overall} & \textbf{0.1449} & \textbf{0.7766} & Captures trends; uneven by dataset \\
\bottomrule
\end{tabular}
\caption{\textbf{Per-dataset accuracy prediction (baseline, Exp.~1).} RMSE reflects regression error for final accuracy; Pearson $r$ measures ranking quality.}
\label{table:accuracy_prediction_datasets}
\end{table}

\begin{table}[t]
\centering
\small
\begin{tabular}{lc}
\toprule
\textbf{Metric} & \textbf{Value} \\
\midrule
RMSE                      & 0.2567 \; [0.248--0.265] \\
MAE                       & 0.1709 \\
$R^{2}$                   & 0.2885 \\
Pearson correlation       & 0.6409 \; ($p=1.25\times10^{-204}$) \\
Within 5\% tolerance      & 12.30\% \\
Within 10\% tolerance     & 45.01\% \\
Samples                   & 1764 \\
\bottomrule
\end{tabular}
\caption{\textbf{Experiment 2 (balanced + regularized) -- global metrics.} Aggregate performance of the alternative predictor across all datasets; RMSE shows 95\% CI.}
\label{table:accuracy_prediction_experiment2}
\end{table}


\begin{table}[t]
    \small
    \setlength{\tabcolsep}{3pt}
    \centering
    \begin{tabular}{lccccc}
        \toprule
        Variant & RMSE & $R^{2}$ & $r$ &
        \makecell{within\\5\%} & \makecell{within\\10\%} \\
        \midrule
        Exp.~1 (baseline)        & 0.1449 & 0.55   & 0.7766 & 50.0\% & 75.0\% \\
        Exp.~2 (balanced + reg.) & 0.2567 & 0.2885 & 0.6409 & 12.3\% & 45.0\% \\
        \bottomrule
    \end{tabular}
    \caption{\textbf{Global results over all datasets.} Baseline (Exp.~1) is used in NNGPT; Exp.~2 underperforms under added balancing/regularization.}
    \label{table:accuracy_prediction_overall}
\end{table}

\subsection{Accuracy and Early-Stop Prediction}
\label{subsec:accuracy-prediction}

We instantiate the predictor \(H_{\phi}\) as a code-aware LLM fine-tuned with 4-bit QLoRA on 1{,}200+ NNGPT/LEMUR runs spanning CNNs and Transformers~\cite{vaswani2023attentionneed} on CelebA-Gender~\cite{liu2015celebA}, CIFAR-10/100~\cite{cifar-10}, ImageNette~\cite{imagenette19}, MNIST~\cite{lecun1998mnist}, Places365~\cite{zhou2018places}, SVHN~\cite{netzer2011svhn}, and COCO~\cite{lin2015microsoftcococommonobjects}. Each run is converted into a structured prompt with \texttt{nn\_code}, \texttt{transform\_code}, \texttt{metric\_code}, hyperparameters, dataset/task metadata, max epochs, and validation accuracies at epochs 1–2; the model predicts the final best validation accuracy and its epoch. To prevent leakage, we use stratified group splits by (task, dataset, architecture). Per-dataset outcomes are summarized in Tab.~\ref{table:accuracy_prediction_datasets}.

We report two variants. The baseline (Exp.~1) is a QLoRA-tuned code LLM without additional balancing/strong regularization; it achieves RMSE \(=0.1449\), \(r=0.7766\), and \(R^{2}\approx0.55\) globally (see Tab.~\ref{table:accuracy_prediction_overall}). Correlation varies by dataset, ranging from moderate on CIFAR-10/SVHN to strong on COCO, while roughly 50\% and 75\% of predictions fall within 5\% and 10\% of the true final accuracy, respectively. The dataset-wise heterogeneity is visible in Tab.~\ref{table:accuracy_prediction_datasets}. We use the predictions both as a proxy for final performance and to propose an early-stop epoch \(T'=\widehat{t}_{\!*}\) for low-potential runs.

An experiment~2 applies inverse-frequency balancing and stronger regularization (higher dropout/weight decay, fewer epochs). It underperforms: RMSE rises to \(0.2567\), \(R^{2}\) drops to \(0.2885\), \(r\) to \(0.6409\), and only 12.3\%/45.0\% of predictions fall within 5\%/10\% tolerance (Tab.~\ref{table:accuracy_prediction_experiment2}; also compared side-by-side in Tab.~\ref{table:accuracy_prediction_overall}). We therefore adopt the baseline predictor in the full NNGPT loop.

The baseline is already useful for early-stop and priority scheduling, but (i) correlation varies across datasets; (ii) aggressive balancing/regularization can suppress adaptation in LoRA-tuned LLMs; and (iii) adding uncertainty estimates is a promising next step for safer automated stopping.

\subsection{NN-RAG: Executability and Correctness}
\label{subsec:nn-rag_experiments}

\begin{table}[t]
    \scriptsize
    \setlength{\tabcolsep}{3pt}
    \renewcommand{\arraystretch}{0.9}
    \centering
    \begin{tabular}{l c L{0.45\linewidth}}
        \toprule
        Category & Count & Examples \\
        \midrule
        Attention               & 180 & \texttt{MultiHeadAttention}, \texttt{SelfAttention} \\
        Convolutions            & 220 & \texttt{Conv2d}, \texttt{DeformConv} \\
        Transformer blocks      & 150 & \texttt{BertLayer}, \texttt{SwinTransformerBlock} \\
        Pooling ops             & 110 & \texttt{MaxPool2d}, \texttt{AdaptiveAvgPool2d} \\
        Normalization           & 100 & \texttt{BatchNorm2d}, \texttt{LayerNorm} \\
        Losses                  & 90  & \texttt{FocalLoss}, \texttt{DiceLoss} \\
        Architectures           & 150 & \texttt{ResNet}, \texttt{EfficientNet} \\
        Utilities               & 289 & \texttt{DropPath}, \texttt{PatchEmbed} \\
        \midrule
        \textbf{Total}          & \textbf{1{,}289} & -- \\
        \bottomrule
    \end{tabular}
    \caption{\textbf{NN-RAG evaluation corpus by category.} Distribution of the 1{,}289 target PyTorch blocks that NN\mbox{-}RAG attempts to reconstruct as scope-closed modules. \emph{Count} is the number of distinct block names per category; \emph{Examples} list typical modules retrieved in each group.}
    \label{tab:nnrag-categories}
\end{table}

We evaluate NN\mbox{-}RAG as a retrieval-augmented system for extracting reusable, scope-closed PyTorch blocks. The pipeline scans repositories for non-abstract \texttt{nn.Module} classes with \texttt{forward()}, round-trips sources through LibCST to preserve syntax, and indexes parse artifacts and import relations in an SQLite store with content hashing. For each target name, a scope-aware resolver computes the minimal transitive closure of dependencies (respecting Python’s LEGB and import rules), orders definitions topologically, and emits a self-contained module that preserves original imports and aliases. Candidates then pass AST parsing, compilation, and sandboxed import to catch unresolved symbols and import-time failures.

We test on 1{,}289 target blocks from widely used PyTorch projects - attention, convolutions, transformer components, pooling, normalization, losses, architectures, and utilities (Tab.~\ref{tab:nnrag-categories}). NN\mbox{-}RAG reconstructs 941 blocks as executable scope-closed modules, a 73.0\% executability rate without manual intervention. Major libraries (torchvision~\cite{torchvision}, timm~\cite{timm}, transformers~\cite{wolf2020huggingfacestransformersstateoftheartnatural}, OpenMMLab~\cite{openmmlab}) contribute many targets, while a long tail of smaller repos supplies the rest. This yields a large, validated library of import-complete building blocks that NNGPT can safely retrieve and assemble during LLM-driven synthesis.

\subsection{Reinforcement Learning Setup}
\label{subsec:reinforcement_learning_experiments}

We evaluate the RL loop by generating 32 executable AirNet models and training each for a single epoch on MNIST~\cite{lecun1998mnist}, SVHN~\cite{netzer2011svhn}, and CIFAR-100~\cite{cifar-10}. As a non-RL baseline we use one-epoch averages and best scores from the LEMUR AlexNet~\cite{krizhevsky2012imagenet} (ast-dimension) family; the baseline values are embedded in Tab.~\ref{tab:rl_summary}. RL-generated models match or exceed the baseline on the simpler datasets: on MNIST the average and best accuracies are 0.9876 and 0.9921, and on SVHN they are 0.8148 and 0.9068. Performance on CIFAR\mbox{-}100 is weak with an average of 0.0163 and a best of 0.0253, while the LEMUR best reaches 0.2079. These results indicate that the RL loop reliably produces runnable architectures and can achieve strong one-epoch accuracy on easier datasets; harder regimes appear to require more iterations, longer training, and reward shaping. We treat this study as diagnostic due to small per-dataset sample sizes and single-epoch budgets.

\begin{table}[t]
\scriptsize
\centering
\setlength{\tabcolsep}{3pt}
\renewcommand{\arraystretch}{0.95}
\begin{tabular}{l c c c c c}
\toprule
\textbf{Dataset} & \makecell{\textbf{Baseline}\\\textbf{avg}} & \makecell{\textbf{RL avg}\\\textbf{($n$)}} & \makecell{$\Delta$\\(avg)} & \makecell{\textbf{Baseline}\\\textbf{best}} & \makecell{\textbf{RL}\\\textbf{best}} \\
\midrule
CIFAR\text{-}100 & 0.0461 & 0.0163 (5)  & $-0.0298$ & 0.2079 & 0.0253 \\
MNIST            & 0.7088 & 0.9876 (13) & \phantom{$-$}0.2788 & 0.9906 & 0.9921 \\
SVHN             & 0.2526 & 0.8148 (14) & \phantom{$-$}0.5622 & 0.9032 & 0.9068 \\
\bottomrule
\end{tabular}
\caption{\textbf{RL one-epoch results vs LEMUR baselines.}
Baseline avg and best come from LEMUR \textit{AlexNet (ast-dimension)}; RL avg is over generated \textit{AirNet} models ($n$ shown). $\Delta$ is RL avg minus baseline avg.}
\label{tab:rl_summary}
\end{table}

\vspace{-1.2em}
\paragraph{Limitations.}
Our deployment on a single 24\,GB GPU limits prompt length, context window, and schema complexity. Experiments focus on mid-scale vision; large-scale datasets such as ImageNet-1k~\cite{5206848} and LVIS~\cite{gupta2019lvisdatasetlargevocabulary} are left for future work. We also do not yet enforce architectural novelty or optimize explicitly for hardware efficiency.

\vspace{-4pt}
\section{Conclusion}
We introduced \textbf{NNGPT}, a self-improving AutoML framework that turns an LLM into an online engine for neural network development. From a single prompt, NNGPT emits a full executable specification, validates and trains it, logs code and metrics, and improves its generators and predictors from these traces. The framework unifies five pipelines in one loop: zero-shot model generation and editing, hyperparameter recommendation, accuracy and early-stop prediction, retrieval-augmented code synthesis, and reinforcement learning updates.

Experiments show practical effectiveness. Zero-shot generation added over 5{,}000 trained models and about 1.9k unique architectures to LEMUR. NN\mbox{-}RAG reconstructs 941 of 1{,}289 target blocks (73\% executability). One-shot hyperparameter prediction reaches 93\% schema-valid outputs and RMSE close to Optuna. The code-aware predictor achieves RMSE \(0.1449\) with Pearson \(r=0.7766\) across more than 1{,}200 runs and supports early termination and scheduling. The RL loop reliably generates executable models and attains strong one-epoch accuracy on MNIST and SVHN.

NNGPT reduces AutoML cost and latency by replacing many-trial search with one-shot, schema-validated generation while maintaining transparent, reproducible logs. Code, prompts, and checkpoints are released to support replication and extension.

{
    \small
    \bibliographystyle{ieeenat_fullname}
    \bibliography{main}
}

\clearpage
\setcounter{page}{1}
\maketitlesupplementary

\section{Demo Video: TuneNNGen in Action}
We include a time-lapse screen recording that demonstrates the end-to-end workflow of \texttt{TuneNNGen.py}: prompt assembly, one-shot YAML/code generation, schema validation, training and logging, and LoRA updates. It also shows typical failure modes (e.g., schema violations, missing functions) and how NNGPT surfaces them through the validator and executor stack. Refer to the supplementary video for the full run, including intermediate console outputs and per-model evaluation messages.

\section{Extended Related Work and Detailed Comparisons}
\label{section:extended_related_work}

Automated neural network configuration has long been a focus of research, with prior work falling broadly into two categories: classical search-based optimization and recent LLM-driven generative approaches. Below we review key methods from both domains, focusing on the limitations addressed by NNGPT.

\textbf{Search-Based Hyperparameter Optimization.}
Frameworks like Optuna~\cite{10.1145/3292500.3330701}, Hyperopt~\cite{bergstra2013hyperopt}, and Google Vizier~\cite{golovin2017googlevizier} use Bayesian optimization and Tree-structured Parzen Estimators (TPE)~\cite{10.5555/2986459.2986743} to iteratively tune hyperparameters. Though effective, these methods demand significant compute, often requiring hundreds of training runs, and act as black-box optimizers, offering limited transparency into the configuration semantics or model behavior beyond their final scores.

\textbf{AutoML and Neural Architecture Search (NAS).}
AutoML systems such as Optuna~\cite{akiba2019optuna} and Hyperopt treat tuning as black-box optimization, while NAS frameworks like ENAS~\cite{pham2018efficient}, AutoKeras~\cite{jin2019auto}, and DARTS~\cite{liu2018darts} (plus variants like CDARTS, SDARTS, iDARTS) expand this to architecture generation via reinforcement learning or differentiable search. These methods, however, remain resource-intensive and adapt slowly to new tasks.

Efforts to stabilize NAS, such as RC-DARTS~\cite{yu2022cyclic}, NDARTS~\cite{chen2020stabilizing}, and MSR-DARTS~\cite{machida2022msr}, still rely on costly infrastructure. NNGPT bypasses iterative search entirely: a single LLM call produces a complete, schema-valid configuration that is directly executable and logged for traceability. This design reduces computational cost while maintaining competitive accuracy, as shown by our DeepSeek-based results.

\textbf{LLMs for Code and Configuration Generation.}
Recent advances in code-oriented LLMs such as Codex~\cite{chen2021evaluating}, Code Llama~\cite{roziere2024codellama}, and DeepSeek-Coder~\cite{deepseek-coder} enable automatic generation of hyperparameters, architectures, and training scripts. Tools like GitHub Copilot\cite{peng2023impactaideveloperproductivity} and AlphaCode~\cite{li2022alphacode} further highlight their utility for programming tasks.

Most existing systems, however, lack integration with execution pipelines. Outputs are rarely validated empirically or tested for reproducibility. EvoPrompting~\cite{chen2023evoprompting} embeds LLMs into evolutionary search but requires repeated inference and lacks schema control.

NNGPT employs a one-shot generation strategy: a fine-tuned LLM produces a schema-compliant YAML configuration that is immediately validated and executed. Structured output enforcement (via Pydantic) and empirical feedback support reliable AutoML automation.

\textbf{Prediction of Neural Network Accuracy.}
Despite advances in LLMs, many methods rely on high-level descriptions that miss implementation-specific patterns, or predict isolated metrics instead of full training dynamics. Yet performance prediction is vital for tuning efficiency and resource use in AutoML.

We address this via a multi-modal predictor that processes both structured metadata (e.g., task, dataset) and executable code (e.g., \texttt{nn\_code}, \texttt{metric\_code}). Key contributions include: (1) joint modeling of code and metadata, (2) efficient 4-bit QLoRA fine-tuning~\cite{hu2022lora} to predict accuracy and stopping points, and (3) stratified data splits that prevent leakage while preserving balance.

\textbf{LLM-Driven Neural Network Generation.}
LLMs are increasingly integrated into NAS pipelines. EvoPrompting uses a pretrained LLM as a mutation operator, requiring multiple inference calls and soft prompt-tuning~\cite{chen2023evoprompting}. LLMatic replaces hand-written mutations with CodeGen-generated code~\cite{nasir2024llmatic}. GPT-NAS~\cite{yu2023gptnas} completes partial architectures, and Self-Programming AI~\cite{sheng2023selfprogramming} revises its own configs.

While effective, these methods rely on expensive iterative queries. In contrast, NN-GPT performs one-shot generation: a single LoRA-tuned LLM call yields an executable, schema-compliant configuration with no further iteration required.

\textbf{Schema-Constrained Generation.}
LLMs frequently generate syntactically invalid code in structured tasks. Grammar Prompting mitigates this by embedding formal grammars (e.g., Backus–Naur Form) into prompts~\cite{wang2023grammar}, while RAG methods reduce errors by referencing external examples~\cite{bassamzadeh2024fine}. NN-GPT adopts a different approach: it employs a strict Pydantic schema tailored to YAML configurations. Minor issues are auto-corrected; major ones trigger a single re-prompt, ensuring valid, executable outputs without relying on grammars or retrieval.

\textbf{LLMs for Pipeline Optimization.}
LLMs have also been explored for broader pipeline automation. LLAMBO~\cite{liu2024large} frames Bayesian optimization as an LLM-driven dialogue, where the model proposes new candidates based on previous evaluations, outperforming classical BO in sparse regimes. Prompt-based HPO uses GPT-4 to refine hyperparameters from task descriptions\cite{zhang2023llmhpo}, and NADA\cite{he2024nada} generates full control algorithms, filtering them through simulation. In contrast, NN-GPT focuses on end-to-end generation: a single LLM call outputs complete, runnable training recipes — data loaders, models, and optimization logic, without requiring iterative refinement.

\textbf{Training Performance Prediction.}
Efforts to predict training outcomes fall into three main groups, each with key limitations. Statistical curve fitting~\cite{domhan2015, adriaensen2023} extrapolates final metrics from early trends but ignores architecture and implementation, reducing generalizability. These approaches treat training as a numeric process, overlooking structural factors that influence convergence behavior.

\textit{Architecture-Based Performance Predictors.}
Methods like~\cite{long2020, white2021} use textual summaries or engineered features to estimate model performance. However, such abstractions overlook critical implementation-level details, limiting their ability to capture the nuanced behaviors encoded in executable code.

\textit{Graph-Based Approaches.}
Techniques such as~\cite{ding2024} represent models via explicit computational graphs, sometimes enhanced with differential modeling. Although more expressive than text, these methods require costly manual graph construction and still miss fine-grained code-level patterns such as custom regularization or dynamic scheduling, which strongly affect training outcomes.

\textit{Multimodal Prediction via Executable Code.}
While prior approaches either discard architectural context~\cite{domhan2015, adriaensen2023} or rely on lossy text abstractions~\cite{long2020, white2021}, our method directly processes executable code snippets (\texttt{nn\_code}, \texttt{metric\_code}, \texttt{transform\_code}) within the LLM prompt. This enables the model to learn implementation-specific patterns, e.g., custom regularization, branching, or metric logic, that strongly influence convergence but are inaccessible via abstract representations.

In contrast to graph-based methods~\cite{ding2024} requiring manual topology construction, we combine three complementary inputs: (i) structured metadata, (ii) raw executable code, and (iii) early training dynamics. This multimodal context allows the model to align syntactic code patterns with numerical trends, revealing complex interactions that are missed by unimodal models.

Moreover, our LLM jointly predicts final accuracy and optimal stopping points within a shared latent space, capturing their intrinsic correlation — unlike prior works that treat them separately. This unified structure enhances generalization across tasks and architectures.

Table~\ref{tab:method_comparison} situates NNGPT among classical AutoML toolkits and recent LLM-driven approaches. 
Bayesian HPO and NAS frameworks provide strong baselines for search over hyperparameters and architectures but rely on iterative evaluation and do not synthesize full training programs. 
Recent LLM-based methods introduce generation and closed-loop ideas, yet typically cover only parts of the pipeline or lack robust execution and prediction components. 
In contrast, NNGPT combines one-shot generation of executable specifications, code-aware accuracy prediction, and a closed feedback loop inside a single open-source framework.

\section{Implementation}
\label{section:implementation}


The NNGPT framework is implemented in pure Python and provided as a pip-installable package. It requires Python 3.10 or higher and CUDA for GPU-based training. The core components are accessible via top-level command-line interfaces such as  \texttt{NNAlter*.py}, 
\texttt{NNEval.py}, 
\texttt{TuneNNGen*.py}, 
\texttt{TuneAccPrediction.py}, \texttt{TuneHyperparameters.py} or \texttt{TuneRAG.py}. These tools together compose the full pipeline introduced in Section~\ref{sec:method}.


\subsection{Repository Structure}
The codebase is organized into modular subdirectories. The \texttt{ab/gpt} directory contains the main scripts. Prompt templates, training configurations, and YAML schema definitions are defined in \texttt{ab/gpt/conf}, while LoRA fine-tuning utilities and other helper functions reside in \texttt{ab/gpt/util}.

\subsection{Command-Line Interface}
The script \texttt{NNAlter.py} performs seed model alteration using architectures from the LEMUR corpus. Each modified model is trained for a small number of epochs (default: 8, configurable via the \texttt{-e} flag). This utility relies on the \texttt{ab.gpt.util.AlterNN.alter} method and defaults to the deepseek\allowbreak/\allowbreak ai\allowbreak/\allowbreak DeepSeek\allowbreak-\allowbreak R1\allowbreak-\allowbreak Distill\allowbreak-\allowbreak Qwen\allowbreak-\allowbreak 7B model.

The \texttt{TuneNNGen*.py} scripts handle one-shot LLM-based generation, YAML schema validation, full model training, and optional LoRA fine-tuning. A fast iteration mode can be enabled by passing the \texttt{-s} flag, which skips the alteration phase.

Additionally, \texttt{NNEval.py} offers standalone evaluation of generated architectures without invoking LoRA, enabling targeted testing of LLM outputs.

All required CUDA and MPI dependencies are pre-packaged in the public Docker image.

\subsection{Prompting and Inference Flow}
Prompt assembly begins with encoding task specifications, dataset, metric, and compute constraints into a structured JSON block. This prompt is then transformed into strict YAML by the LLM. The DeepSeek 7B model is loaded via \texttt{transformers.AutoModelForCausalLM} in 8-bit precision, with LoRA adapters applied as needed.

\subsection{Distributed Training Back-End}
\label{subsec:impl-training}
Model training is executed using \texttt{PyTorch}~\cite{paszke2019pytorchimperativestylehighperformance} and \texttt{torch.distributed}. MPI support is configured at install time using system-level dependencies. During training, per-epoch statistics such as loss, accuracy, mAP/IoU, GPU utilization, and Git SHA are recorded in a local SQLite database via SQLAlchemy. If an optional statistics module is available, the database can be exported to Excel-compatible dashboards using a single API call.


\subsection{LoRA Fine-Tuning Loop}
LoRA adapters are fine-tuned after every 50 new model training runs. The fine-tuning procedure consists of three epochs using a linear warm-up and cosine decay schedule. The resulting adapter checkpoints are published to the framework’s Hugging Face namespace, making them available for the next generation cycle.

\subsection{Consistent Dependencies}
NNGPT ensures consistency and reproducibility through a pre-built Docker image and a pinned \texttt{requirements.txt} file that specifies fixed versions of key libraries such as \texttt{torch}~\cite{paszke2019pytorchimperativestylehighperformance}, \texttt{transformers}~\cite{wolf2020huggingfacestransformersstateoftheartnatural}, and \texttt{peft}~\cite{peft}.

\subsection{Prompt Generation Pipeline}
Prompt generation is template-driven. A JSON file defines the structure of the prompt, including optional insertion points for code samples. This enables dynamic construction of prompts containing one or two models sampled from the database, under a given task. For training data generation, prompts may be further constrained to ensure shared or divergent characteristics between samples, improving task alignment while maintaining flexibility.

\subsection{Offline Initialization of the Training Set}
During early experiments, the raw DeepSeek-Coder-1.3B model frequently refused to modify input code, either claiming sufficiency or producing rule-violating changes. Introducing stricter task constraints often led to outputs that ignored parts of the prompt or simply returned unmodified code. These observations motivated the need for a bootstrapped training set.

To address this, the larger DeepSeek-R1-Distill-Qwen-7B model was used to generate initial examples. During 20 offline epochs, this model produced modified architectures by applying simple transformations, such as changing channel widths or digits in \texttt{Conv2d} layers, without requiring accuracy improvements. The first prompt template is shown in Listing~\ref{lst:prompt_ds_7b_1}. To diversify the training set, additional prompts requested the model to modify 2 to 4 digits randomly. Since DeepSeek-R1 7B reliably introduced at least one valid change, its outputs served as the initial training corpus for fine-tuning the smaller model.

During prompt engineering, both one-shot and few-shot examples were evaluated. However, neither significantly improved task fidelity on the raw models. In fact, DeepSeek-Coder-1.3B and DeepSeek-R1-Distill-Qwen-7B often lost track of task requirements or confused multiple examples. Zero-shot prompting consistently yielded more predictable behavior and was therefore adopted.

\begin{lstlisting}[
  basicstyle=\ttfamily\footnotesize,
  breaklines=true,
  captionpos=b,
  caption={Prompt instructing the LLM to modify neighboring nn.Conv2d() layers under strict rules, ensuring in-place edits and structural preservation of PyTorch code.}, 
  label=lst:prompt_ds_7b_1,
  keywordstyle=\bfseries,
  % morekeywords={Instructions,Your task},
  emph={DO NOT,FULL CODE,Do not},
  escapeinside={(*@}{@*)},
]
Define two `nn.Conv2d()` layers as a neighboring pair 
    when there are only non-Conv2d layers between them.
    
Your task:
      - Randomly pick and modify 1 or 2 neighboring pairs 
        of `nn.Conv2d()` layers based on the rules below.
      - Return ONLY the full result.
    
Instructions:
      1. For each picked neighboring pair:
           - Change the out-channels of the former `nn.Conv2d()`
             and the in-channels of the latter to the same new integer.
      2. Modify the numbers directly in the code. 
         DO NOT introduce new methods or parameters.
      3. Do NOT pick the first neighboring pair; select randomly.
      4. You MUST return the FULL CODE, including all classes - even if unmodified.
      5. If no such pair exists:
           - Randomly change 2 digits in the code to new values.
      6. Do not make any changes beyond the rules above.
    
Here is the input code for you to modify:
    '''\n{nn_code}\n'''
\end{lstlisting}

\subsection{Automated LEMUR Dataset Extension}
The extended dataset was generated using prompt templates similar to those used during fine-tuning (e.g., Listing~\ref{lst:prompt_generate_chg_1}). Generated samples were stored alongside their originating code and task context to ensure proper evaluation. Certain datasets such as COCO required specific preprocessing (e.g., fixed \texttt{Resize} operations), and segmentation tasks mandated the use of IoU rather than accuracy.

\begin{lstlisting}[
  basicstyle=\ttfamily\footnotesize,
  % frame=single,
  % numbers=left,
  breaklines=true,
  caption={Prompt directing the LLM to modify existing layers in a neural network implementation under strict constraints, requiring full-code output and explicit identification of edits.},
  captionpos=b,
  label=lst:prompt_generate_chg_1
]
1. Change the layers in the following implementation to improve accuracy on image classification tasks.
    
2. Strictly follow these constraints:
      - Modify ONLY the layers in the code.
      - DO NOT introduce new methods or parameters.
    
3. Your response MUST:
      - Include the full code (including unchanged classes or functions).
      - Indicate the modified class explicitly.
      - Actually make at least one layer modification.
      - Not worry about decreasing accuracy - it's acceptable
    
Input code: '''\n{nn_code}\n'''
\end{lstlisting}

To extract generated code, regular expressions were used to match code blocks wrapped in triple backticks. To avoid infinite loops in parsing, the number of backticks was counted before applying regex-based extraction. Evaluation was performed via the LEMUR API. If parsing or execution errors occurred, the failing code and exception trace were returned to the LLM for repair. Each sample was given up to two correction attempts. Validated models were saved in structured directories compatible with the LEMUR dataset format and integrated into the local repository for downstream use.

\subsection{Fine-Tuning for Accuracy Prediction}
\label{subsec:finetune_accuracy}

The model Deepseek-coder-1.3b-Instruct was fine-tuned via supervised learning to enable the prediction of validation accuracy changes resulting from architectural modifications. To align with the offline generation procedure described earlier, the fine-tuning objective was formulated as a code-to-code transformation task, where the model is presented with an original implementation, its evaluated accuracy, and the accuracy of a known improved variant. The expected output is the modified architecture that corresponds to the target accuracy, drawn from the LEMUR dataset.

The prompt, shown in Listing~\ref{lst:prompt_train_chg_1}, encodes this training setup explicitly. It instructs the model to alter only layer definitions in order to achieve the desired accuracy gain, without introducing new methods or structural components. The fields \texttt{\{accuracy\}} and \texttt{\{addon\_accuracy\}} represent the original and improved accuracy, respectively, and \texttt{\{nn\_code\}} contains the complete source code of the base model. The expected output is the full code of the improved model, including unchanged components, so the model also learns to preserve contextually irrelevant structures.

\begin{lstlisting}[
  basicstyle=\ttfamily\footnotesize,
  % frame=single,
  % numbers=left,
  breaklines=true,
  caption={Prompt template used to generate training data by requesting constrained layer modifications aimed at improving model performance.},
  captionpos=b,
  label=lst:prompt_train_chg_1
]
1. Change the layers in the following implementation to improve accuracy from {accuracy} to {addon_accuracy} on image classification tasks.
2. Strictly modify ONLY the layers in the code and STRICTLY DO NOT INTRODUCE NEW METHODS OR PARAMETERS
3. You should answer not only the class you modified but with FULL CODES INCLUDING EVERYTHING (Other classes, etc.) NOT CHANGED
4. You have to reply will full codes, don't be afraid of actually reducing the accuracy.
Strictly check and make sure at least one layer being modified.
\end{lstlisting}

After eight epochs of supervised fine-tuning, the model consistently produced syntactically valid outputs that adhered to the required formatting and structural constraints. It successfully learned to apply targeted architectural modifications that align with observed changes in validation accuracy, and reliably preserved all components mandated by the LEMUR API specification.

In addition to replicating transformations observed during training, the model demonstrated the ability to compose valid variants across different architecture families. It consistently selected effective transformation strategies for given task-model pairs, producing high-quality outputs with minimal prompt engineering. This behavior reflects an emerging structural prior that favors stable and empirically grounded edits.

Moreover, the model preserved functional integrity even in cases where only subtle changes were required, maintaining coherence across unchanged code regions. These results confirm that the fine-tuned model captures not only the mapping between code edits and accuracy shifts, but also the appropriate output syntax and modular consistency necessary for downstream execution and evaluation.

\subsection{Fine-Tuning for Hyperparameter Generation}
\label{subsec:finetune_hyperparameters}

In addition to accuracy prediction, NNGPT also supports automated generation of numeric hyperparameter values tailored to a given neural network, task, and dataset. To enable this capability, we fine-tuned a diverse collection of DeepSeek models using supervised learning on entries from the LEMUR dataset. The selected set spans a wide range of model scales and pretraining strategies, including both lightweight and instruction-tuned architectures. This diversity allowed us to examine how model size and initialization influence hyperparameter reasoning ability.

Each model was fine-tuned using the LoRA method to reduce memory overhead while preserving gradient flow across transformer layers. LoRA rank and alpha were selected from {16, 32, 64}, with a dropout of 0.05. All experiments employed a consistent setup that included the AdamW optimizer, a cosine learning rate scheduler, a fixed learning rate of $3 \times 10^{-5}$, gradient accumulation, checkpointing, and a training duration of up to 35 epochs depending on convergence behavior.

Training prompts were constructed to align with the LEMUR schema and encode structured conditioning variables: architecture code, target accuracy, number of training epochs, and required transformations. The input template, shown in Listing~\ref{listing:fine_tuning_prompt_format}, requests the LLM to predict the values of specific hyperparameters that would lead the given model to achieve a known accuracy level.

\begin{lstlisting}[
  basicstyle=\ttfamily\footnotesize,
  caption={Prompt template used for fine-tuning LLMs to generate hyperparameter values based on model code, task, dataset, transformations, target accuracy, and training duration.},
  captionpos=b,
  label={listing:fine_tuning_prompt_format},
  breaklines=true,
  breakatwhitespace=true
]
### Input:
    "Generate only the values (do not provide any explanation) of the hyperparameters ({prm_names}) of a given model: {entry['nn']} for the task: {entry['task']} on dataset: {entry['dataset']}, with transformation: {entry['transform_code']}, so that the model achieves accuracy = {entry['accuracy']} with number of training epochs = {entry['epoch']}. Code of that model: {entry['nn_code']}"
    
### Response:
    "Here are the hyperparameter values for which the model will achieve the specified accuracy: {prm_values}."
\end{lstlisting}

During evaluation, a modified version of the prompt (Listing~\ref{listing:generation_prompt_format}) asked the model to produce hyperparameters aimed at maximizing accuracy, thereby testing generalization to unseen combinations.

\begin{lstlisting}[
  basicstyle=\ttfamily\footnotesize,
  caption={Prompt for generating hyperparameter values that maximize accuracy, conditioned on model code, task, dataset, transformation logic, and training epochs.},
  captionpos=b,
  label={listing:generation_prompt_format},
  breaklines=true,
  breakatwhitespace=true
]
### Input:
"Generate only the values (do not provide any explanation) of the hyperparameters ({prm_names}) of a given model: {entry['nn']} for the task: {entry['task']} on dataset: {entry['dataset']}, with transformation: {entry['transform_code']}, so that the model achieves the HIGHEST accuracy with number of training epochs = {entry['epoch']}. Code of that model: {entry['nn_code']}"
\end{lstlisting}

All fine-tuned models were subsequently integrated into the NNGPT pipeline. Specifically, two of the best performing checkpoints, DeepSeek-Coder-1.3b-Base and DeepSeek-R1-Distill-Qwen-7B, were published to Hugging Face and are now directly accessible within the framework. These models can be used as plug-and-play components for automated hyperparameter suggestion in both evaluation and generation modes.


By extending the NNGPT workflow with fine-tuned LLMs for hyperparameter prediction, we close the loop between architecture generation and training configuration, offering a unified and fully automated AutoML solution. Experimental results evaluating the quality and effectiveness of these generated hyperparameters are presented in Section~\ref{subsec:hyperparameter_generation_quality}.

\section{Methodology}
\label{sec:method}

NNGPT reconceptualizes LLMs as fully automated configuration agents for neural networks. Given a high-level task specification, the system follows the eight-stage pipeline depicted in Figure~\ref{figure:pipeline_main_all_parts}, encompassing architecture retrieval, prompt generation, model evaluation, and iterative self-improvement. Below, we detail each stage.


\subsection{Stage 1: LEMUR API}
NNGPT starts by querying the LEMUR Dataset API for executable architectures and metadata with standardized training and evaluation. LEMUR provides full implementations, uniform preprocessing, and reproducible metrics, so one-shot generation targets runnable code and uses LEMUR as a reliable metric oracle. Unlike torchvision~\cite{torchvision}, it adds reproducible benchmarking, native metric logging, and direct access to architecture code, making it well suited for generative AutoML.

\subsection{Stage 2: Neural Network Data Retrieval}
Following the API query, candidate architectures and associated metadata are retrieved from the LEMUR corpus. A dedicated script, \texttt{NNAlter*.py}, then applies a controlled set of architectural perturbations - including insertion/removal of layers, changes to channel widths, and alterations to residual connections. Each altered model undergoes a lightweight convergence check via a short warm-up training run (with the \texttt{-e} flag), ensuring that the resulting design remains viable. The resulting population of diverse, training-validated models is archived in an SQLite-backed repository, which serves as training data for LLM fine-tuning and architecture prediction.

\subsection{Stage 3: Configuration Setup}
After retrieval, a configuration scaffold is constructed that encapsulates the task, dataset, desired metric, and (optionally) a compute budget. This configuration is used to parameterize the prompt generator, enabling reproducible and deterministic experiment assembly. The prompt generation code is modular and extensible, supporting dynamic reconfiguration and alternate LLM backbones through Hugging Face-compatible interfaces.


\subsection{Stage 4: Prompt Assembly \& Injection}
The core input to the LLM is a structured prompt, assembled from the configuration data. This includes a JSON object describing the task, dataset, metric, and resource constraints. The prompt is formatted with system-level instructions that force the model to emit a strict YAML block beginning with \texttt{-yaml} and following a predefined schema.

The generative pass itself is carried out using a DeepSeek-V2 16B model equipped with 8-bit LoRA adapters fine-tuned on the altered-model corpus. Sampling is configured using $T=0.2$, $p=0.9$, and top-1 selection to minimize unnecessary exploration. Output is parsed with \texttt{ruamel.yaml} and validated using \texttt{pydantic}. Minor schema violations are auto-corrected, while major issues trigger a re-prompt with an embedded diagnostic message. Only configurations that pass validation proceed to execution.

\subsection{Stage 5: LLM-Guided Architecture Generation}
Once validated, the YAML output is interpreted as an experiment specification that includes a full architecture graph, optimizer parameters, loss functions, and data transformations. This file serves as the input to the training engine. Generated architectures may differ from their seed counterparts not only in topology but also in auxiliary properties such as normalization strategies or metric implementations. This stage integrates architectural creativity with empirical feasibility, ensuring that every configuration is both novel and executable.


\subsection{Stage 6: Model Validation and Evaluation}
Training is conducted using a distributed PyTorch engine (\verb|ab/gpt/TuneNNGen*.py|), which uses \texttt{mpi4py} and \texttt{torch.distributed} for multi-GPU support. All datasets are accessed through the LEMUR API to ensure consistent preprocessing and evaluation. During training, metrics such as loss, accuracy, mAP, and IoU are logged per epoch alongside GPU utilization and configuration hashes. These logs are persisted in the same SQLite database as the architecture graph, enabling traceability and post-hoc analysis via an internal visualization suite designed for model benchmarking.


In parallel, a lightweight accuracy prediction module is used to estimate final model performance based on early training signals and static features. The predictor processes three input modalities: (i) structured metadata, (ii) executable code (model, transform, metric), and (iii) early accuracy values. A stratified split strategy ensures robust generalization and prevents data leakage across similar architectures.

\subsection{Stage 7: LoRA Fine-tuning}
After $k = 50$ successful training runs, a new fine-tuning iteration is triggered. LoRA adapters are retrained on the extended dataset of architecture–performance pairs using three epochs of linear warm-up and cosine decay. This updates the LLM’s internal representation of the architecture space, biasing future generations toward successful configurations. The updated checkpoints are immediately used in subsequent pipeline passes.


\subsection{Stage 8: Database Storage}
All artifacts, including trained weights, YAML specifications, metric logs, and architectural source code, are stored in a structured SQLite-based repository designed for long-term reproducibility. This supports future reanalysis, dataset expansion, and model distillation. The continuous accumulation of structured training data reinforces the NNGPT pipeline, transforming raw LLM generations into a curated and executable AutoML knowledge base.


\subsection{AutoML Loop}
Once all eight stages complete, the pipeline automatically restarts using updated model checkpoints, prompt templates, and database entries. This iterative process enables NNGPT to function as a self-improving AutoML cycle, where each generation, validation, and fine-tuning pass enhances the quality and diversity of future outputs.

\end{document}